\documentclass[letterpaper, 10 pt, conference]{ieeeconf}
\IEEEoverridecommandlockouts
\usepackage{cite}
\usepackage{amsmath,amssymb,amsfonts}
\usepackage{algorithmic}
\usepackage{graphicx} 
\usepackage{hyperref}
\usepackage{color}
\usepackage{comment}
\usepackage{xcolor}
\usepackage{subcaption}
\usepackage{float}
\usepackage{diagbox}
\def\BibTeX{{\rm B\kern-.05em{\sc i\kern-.025em b}\kern-.08em
    T\kern-.1667em\lower.7ex\hbox{E}\kern-.125emX}}

\begin{document}

\title{\LARGE \bf Solving Stochastic Orienteering Problems with Chance Constraints Using a GNN Powered Monte Carlo Tree Search}

\author{Marcos Abel Zuzu\'{a}rregui \qquad Stefano Carpin%
\thanks{
The authors are with the Department of Computer Science and Engineering, University of California, Merced, CA, USA.
This work is partially supported by 
is partially supported by USDA-NIFA under award \# 2021-67022-33452 (National Robotics Initiative) and by the IoT4Ag Engineering Research Center funded by the National Science Foundation (NSF) under NSF Cooperative Agreement Number EEC-1941529. Any opinions, findings, conclusions, or recommendations expressed in this publication are those of the author(s) and do not necessarily reflect the view of the U.S. Department of Agriculture or the National Science Foundation.}}%

\maketitle

\begin{abstract} 
Leveraging the power of a graph neural network (GNN) with message passing, we present a Monte Carlo Tree Search (MCTS) method to solve stochastic orienteering problems with chance constraints. While adhering to an assigned travel budget, the algorithm seeks to maximize collected reward while incurring stochastic travel costs. In this context, the acceptable probability of exceeding the assigned budget is expressed as a chance constraint. Our MCTS solution is an online and anytime algorithm, alternating planning and execution, that determines the next vertex to visit by continuously monitoring the remaining travel budget. The novelty of our work is that the rollout phase in the MCTS framework is implemented using a message-passing GNN, predicting both the utility and failure probability of each available action. This allows to enormously expedite the search process. Our experimental evaluation shows that with the proposed method and architecture, we manage to efficiently solve complex problem instances while incurring moderate losses in terms of collected reward. Moreover, we demonstrate how the approach is capable of generalizing beyond the characteristics of the training dataset. The paper’s website, open-source code, and supplementary documentation can be found at \url{ucmercedrobotics.github.io/gnn-sop}.
\end{abstract}

\section{Introduction}

\begin{figure}[htb]
    \centering
    \includegraphics[width=0.6\linewidth]{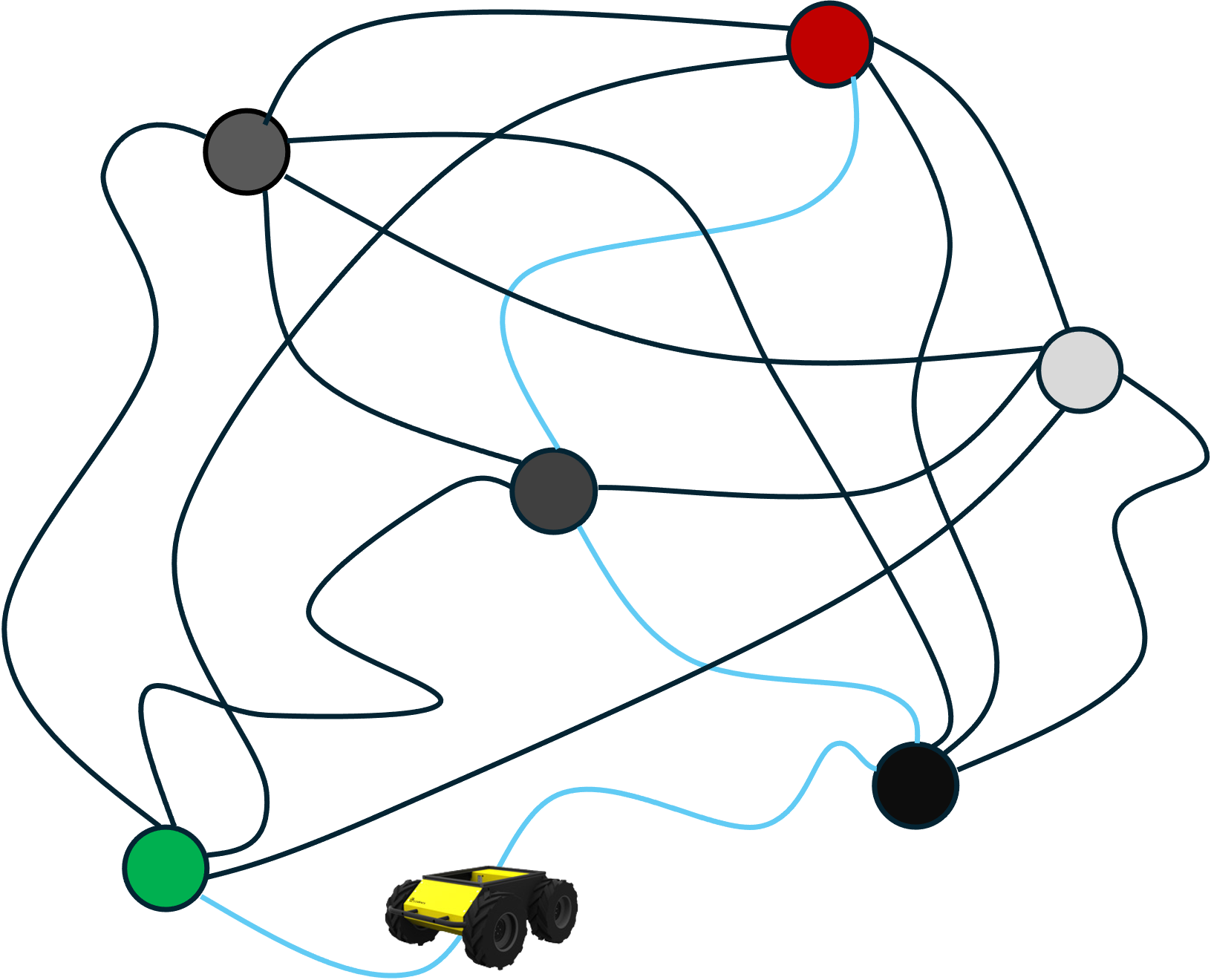}
    \caption{
    The orienteering problem is defined on a graph $G$ where vertices have rewards and edges have non-negative costs. The goal is to find a path from an assigned start vertex (the red node in this example) to a goal vertex (the green node) that maximizes the sum of collected rewards while ensuring the path cost does not exceed a given budget, $B$. 
    }
    \label{fig:sop}
        \vspace{-3mm}
\end{figure}

Orienteering is an APX-hard optimization problem defined on a weighted graph, $G$, where each vertex has a reward and each edge has a non-negative cost \cite{Golden1987}.
The goal is to plan a path between designated start and end vertices to maximize the total reward collected from visited vertices while staying within a budget, $B$, that limits the total path length.
This budget is often considered in terms of time, power, or distance that can be traveled (see Figure \ref{fig:sop}).
Unlike the Traveling Salesman Problem (TSP), a typical solution to an orienteering problem instance does not visit all nodes.
Instances of the orienteering problem can model various real-world scenarios we encounter daily, such as logistics \cite{CarpinTRO2022}, surveillance \cite{yu2014correlated, 10008951}, ridesharing \cite{Martin2021}, and precision agriculture \cite{CarpinTASE2020}, among others.
Our interest in this problem is driven by its applications in precision agriculture \cite{CarpinIROS2020,CarpinRAL2021,CarpinIROS2021,CarpinTASE2024}, though its range of uses is broad and continually expanding.
Most research on  orienteering  has concentrated on the deterministic version, where both vertex rewards and edge costs are known in advance. However, this is not often the case in many practical scenarios. For example, when a robot moves from one location to another, the time or energy required for the movement is often a random variable rather than a deterministic value. This variant is known as the \emph{stochastic orienteering problem} (SOP). Although SOP has been studied \cite{Campbell2011}, it has received significantly less attention than its deterministic counterpart.

Recently, we proposed a novel solution to the stochastic orienteering problem
using Monte Carlo Tree Search (MCTS), aimed at estimating both the collected
reward and the probability of exceeding the allocated budget $B$
\cite{CarpinTASE2024}. Previous work highlighted in \cite{9109309} and confirmed
through our own experiments demonstrates that the rollout phase of MCTS can
consume up to 99\% of the total planning time.
Aiming to accelerate the search with the goal of
deploying a planner in applications requiring real-time response,
in this work we investigate the potential of using a graph neural network (GNN)
architecture to train a model capable of predicting rollout outcomes. A key
innovation of our approach lies in simultaneously predicting
two critical metrics needed to solve SOP instances: the
accumulated reward and the probability of budget constraint violations. To the
best of our knowledge, this dual prediction has not been explored previously.

Accordingly, in this contribution we present our GNN-powered MCTS SOP solver. 
Our novel contributions as follows:
\begin{itemize}
    \item we introduce a single architecture capable of predicting values
    for a custom developed rollout procedure that can output
    action quality and failure probability;
    \item we demonstrate that this approach is competitive with the model that trained it and even optimal solutions, at times;
    \item we demonstrate that we can plan and execute complex orienteering solutions at real-time speeds.
\end{itemize}
  
The rest of the paper is organized as follows. Selected related work
is presented in Section \ref{sec:sota}. SOP is formalized in
Section \ref{sec:statement}, and in the same section we briefly define the model that generates our training data.
The core of our contribution is in Section \ref{sec:method}, where we describe
the network we developed, as well as the feature attributes we identified to
effectively train the networks.
Extensive simulations
detailing our findings are given in Section \ref{sec:results},
with conclusions and future work discussed in Section \ref{sec:conclusions}.
\section{Related Literature} \label{sec:sota}
Early approaches to solve deterministic and stochastic orienteering
problems relied on classical methods, often using heuristics due to the
intrinsic computational complexity \cite{Gunawan2016}.
Notably, in \cite{MILP} the authors propose an exact solution
based on a mixed-integer linear program formulation to solve the
stochastic orienteering problem with chance constraints.
This approach is however offline, i.e., it computes a solution before
execution starts and does not adapt on the fly.
Our recent work \cite{CarpinTASE2024} instead solves the problem using MCTS, designed to be an online method, whereby the vertices
to visit are selected at runtime based on the remaining budget.
This method, while competitive, devotes a significant amount of time
to the rollout phase in MCTS search.

The practice of embedding policy and/or value networks into MCTS has
been notably used in \cite{alphago} and in subsequent papers such as \cite{swiechowski2018improving}.
Even though these papers do not use GNNs, they provide evidence of being able to solve computationally expensive problems by reducing the simulation required to approximate a solution.
They leverage the power of MCTS and approximate their rollout phases, among other improvements.
Several recent approaches have explored the use of GNN-based
solutions to solve NP-hard graph optimization problems.
For example, \cite{DBLP:journals/corr/DaiKZDS17, 9109309} focused on using GNNs to solve TSP and other graph optimization problems.
While different from orienteering, solutions to TSP are of interest
due to some similarities in the problem formulation.
Both of these papers used a message-passing framework with varying layer configurations.
Importantly, \cite{9109309} used MCTS to explore a limited horizon rather than the entire state space.
In \cite{LIU202446}, the authors show that a similar implementation can be used to solve orienteering problems.
In particular, a GNN-based approach is used
in combination with a beam search method, as done in \cite{9109309}.
However, none of these approaches address stochastic orienteering.
Equally relevant, to the best of our knowledge, no methods combining GNNs and MCTS have been developed to solve the SOP with chance constraints, which is the focus
of this contribution.

\section{Problem Statement and Background}
\label{sec:statement}

In this section, we formally introduce the stochastic orienteering problem with chance constraints (SOPCC). We then outline our previous solution utilizing MCTS, which motivates the method proposed in this manuscript.

    \subsection{Stochastic Orienteering Problem with Chance Constraints}
    The deterministic orienteering problem is defined as follows. Let $G=(V,E)$
    be a weighted graph with $n$ vertices, where $V$ is the set of vertices
    and $E$ is the set of edges. Without loss of generality, we assume
    $G$ is a complete graph. Define $r:V \rightarrow \mathbb{R}$ as the reward function assigning a reward to each vertex, and  $c: E\rightarrow \mathbb{R}_{+}$ as the cost function assigning a positive cost to each edge. Let $v_s,v_g \in V$
    be the designated start and end vertices, respectively, and let $B > 0$ be a fixed budget.
    Note that we allow $v_s=v_g$ for cases where the start and end vertices are the same.
    For a path $P$ in $G$, define $R(P)$ as the sum of the rewards of the vertices
    along $P$, and  $C(P)$ as the sum of the costs
    of the edges in $P$. The deterministic orienteering problem  aims to find
    \[
    P^* = \arg \max_{P \in \Pi}R(P) \qquad \textrm{s.t.}~ ~C(P^*) \leq B
    \]
    where $\Pi$ is the set of simple paths in $G$ starting at $v_s$ and ending at $v_g$. A simple path is defined as a path that does not revisit any vertex. Given the connectivity assumption of $G$, restricting $\Pi$ to 
    simple paths is not limiting.
    In the stochastic version of the problem, the cost associated with each edge is not fixed but rather sampled from a continuous random variable with a known probability density function that has strictly positive support. Specifically, for each edge $e \in E$ we assume $c(e)$ is sampled from $d(e)$ where $d(e)$
    represents the random variable modeling the cost of traversing edge $e$. In general, different edges are associated with different random variables.
    In the stochastic case then, for a given path $P$ the path cost $C(P)$ is also 
    a random variable. The constraint on the path cost $C(P)$ must therefore be expressed using a chance constraint, defined formally as follows.

    \begin{quote} {\bf Stochastic Orienteering Problem with Chance Constraints
    (SOPCC)} With the notation introduced above, let $0< P_f < 1$ be an
    assigned failure bound. The SOPCC seeks to solve the following optimization 
    problem:
     \begin{align}
    P^* & = \arg \max_{P \in \Pi}R(P) \nonumber \\
    \textrm{s.t.}~ &\Pr[C(P^*)>B]\leq P_f.\nonumber
    \end{align}
    \end{quote}

\emph{Remark:} Since the probability density functions of the random variables associated with the edges are assumed to be known, generating random samples for the cost 
$C(P)$ of a path  $P$ is straightforward and does not necessitate computing the probability density function of $C(P)$.
This is consistent with the assumptions usually made when using MCTS to solve
planning problems \cite{gupta_running_2015}.

The problem definition models a decision maker aiming  at maximizing the reward collected along a path while remaining below the allotted budget. Because the 
cost of the path $C(P)$ is a random variable, the constraint can be
satisfied only in probability, and this leads to the introduction of the 
chance constraint $\Pr[C(P^*)>B]\leq P_f.$

\subsection{Monte Carlo Tree Search for SOPCC} \label{MCTS}
In \cite{CarpinTASE2024}, we introduced a new approach based on MCTS
to solve the SOPCC. Key to our new solution was the introduction
of a novel criterion for tree search dubbed UCTF -- Upper Confidence Bound for Tree with Failures. In this subsection, we briefly review this
solving method. We refer the reader to \cite{CarpinTASE2024} for a deeper discussion
of our method and to \cite{coulom2007efficient} for a general introduction to the MCTS methodology.
Note that in \cite{CarpinTASE2024},
when deciding which vertex to add to the route, MCTS only
considers the 
$K$-nearest neighbors to the current vertex. In the version implemented in this paper to train the GNN described in the next section, we remove this feature and instead consider all vertices. The reason
is that in \cite{CarpinTASE2024} we reduced the number of possibilities to
expedite the online search, while here training is done offline. We can therefore consider a larger search space.
As pointed out in \cite{SuttonRL} (chapter 8), a solution
based on MCTS relies on the definition of four steps: selection, expansion, rollout, and backup. In the following, we sketch how these can be
customized for solving SOPCC.

\subsubsection{Selection}
Selection, also known as the \emph{tree policy}, guides the search from the root
to a leaf node based on the value associated with the nodes in the tree.
At every level, each child of the current node is evaluated using a metric,
and the search then proceeds to the node with the highest value.
Universal Confidence Bound for Trees (UCT) is commonly used
to attribute values to nodes \cite{UCB} and defines
the tree policy. UCT, however, aims at maximizing
reward only and is therefore not suitable for use in problems with
chance constraints where high reward could lead to violations of the constraint.
To overcome this limit, in \cite{CarpinTASE2024} we introduced UCTF, defined 
as follows

    \begin{equation}
        \begin{gathered}
        UCTF(v_j) = Q[v_j](1 - F[v_j]) + z \sqrt{\frac{\text{log}(N[v_{p}])}{N[v_j]}}
        \end{gathered}
        \label{eq:UCTF}
    \end{equation}
where the term $(1 - F[v_j])$ is new. 
In Eq.~\eqref{eq:UCTF}, $Q[v_i]$ is the estimate of
how much reward will be collected by adding $v_j$ to the route, $F[v_j]$  is a failure probability 
estimate for a path going through $v_j$, $N[v_j]$ is the number of times
$v_j$ has been explored, and $N[v_{p}]$ is the number of times
its parent has been explored. Critically, in our original approach both $Q$ and 
$F$ are estimated via rollout, while in this paper we will train a GNN
to quickly predict these values. The term $(1 - F[v_j])$ penalizes nodes with 
high estimates for the failure value. As in the classic MCTS method, the term
$z$ balances exploration  with exploitation.

\subsubsection{Expansion}

Expansion is the process of adding a node to the tree. In our implementation,
this is implicitly obtained by  assigning  a UCTF value of infinity to 
vertices that are still unexplored (i.e., for which $N[v_j]$ is 0).
This way, we  ensure all children nodes of a parent node
are explored at least once before one of the siblings is selected again.

\subsubsection{Rollout}

    Once a node is added, rollout (also called simulation) 
    is used to estimate the amount of
    reward that will be collected by going through that node as well
    as its failure probability. This is done by running a baseline, handcrafted, heuristic policy multiple times. The $F$ value is
    estimated as the ratio of failed runs over total runs of the
    heuristic policy, while the $Q$ value is estimated as the average
    of the returned value limited to the successful runs. A critical
    aspect of this step   is the necessity of running the
    heuristic multiple times to numerically estimate $F$.
 In \cite{CarpinTASE2024}, the heuristic is a mix of 
    random exploration and greedy search.

\subsubsection{Backup} \label{backprop}
 The backup step in \cite{CarpinTASE2024} is unique because it propagates
 back not only the $Q$ value, but also the $F$ value. This is because for each
 action we must estimate not only the anticipated reward, $Q$, but also the
 probability, $F$, that the action will eventually result in a budget violation.
  To this end, two cases are considered by comparing the $F$ value of the
  parent with the $F$ value returned by the rollout.
 If the $F$ value for the parent node $v_{p}$ is $\leq P_f$, we check if 
 the $F$ value of the newly expanded node is also below $P_f$. If it is, we then compare the reward estimate, $Q$. 
    If the child node has a larger estimate when adding the reward of the parent node, we update the parent estimate. If not, we move on.
    In the case where the parent node has an $F > P_f$, we check if the child node $F$ estimate is less than the parent's. 
    If it is, we update the parent's $F$ and $Q$ with $F$ being the child's $F$ value and $Q$ being the child's $Q$ plus the true, known reward of the parent node (see \cite{CarpinTASE2024} for additional details).
\section{Methodology}
\label{sec:method}

\subsection{MCTS Value and Failure Network}
The main novelty of the method we propose here is a single value/failure network architecture
that can be trained to predict either expected value, $Q$, or failure probability, $F$, of an
action, $v_j$, to be used in the tree policy UCTF formula from Eq.~\eqref{eq:UCTF}.
Therefore, after appropriate training, the MCTS rollout phase with 
multiple simulations to estimate both $Q$ and $F$ is replaced by 
a single forward pass on each network, 
thus dramatically cutting the computation  time. As a result, as explained later, when a node is considered for expansion,
all descendants are generated and evaluated at once. This is different
from the classic MCTS approach where node descendants are added
and evaluated one by one. This feature greatly extends the set of
actions explored while not requiring more computational time.

\subsection{Message Passing Neural Network}

Given that orienteering problem instances are defined over graphs,
a message passing neural network (MPNN) 
\cite{WuArxivGNNSurvey,ZhouArxivGNNSurvey} is the natural 
choice to implement the value and failure networks.
MPNNs are graph neural networks in which, at every 
iteration of learning, nodes of the graph share information with 
neighboring nodes to generate a $D$-dimensional embedding for each node
or for the entire graph.
In a typical message passing framework, the message update function encompasses node 
attributes as well as edge attributes, 
both of which are critical in the SOPCC problem space.
The  following equations show the general relationship between an embedding and a message within the framework \cite{gilmer2017neural}, formulated in terms 
of the aggregation operation given by Eq.~\eqref{eq:mpnn} (where the set $\mathcal{N}(v)$
is the set of nodes neighboring $v$) and the update operation 
given by Eq.~\eqref{eq:update}.
\begin{align}
    m_v^{t+1} &= \sum_{w \in N(v)}M_t (h_v^t, h^t_w, e_{vw}) \label{eq:mpnn}\\
    h_v^{t+1} &= U_t(h_v^t, m_v^{t+1}) \label{eq:update}
\end{align}

Aggregation defines that a message $m_v^{t+1}$ for node $v$ at time $t+1$ is obtained by
combining the embedding $h_v^t$ with all of its neighbors' embeddings, $h_w^t$, 
at time $t$, as well as the attributes $e_{vw}$ of the edges connecting them.
The update operation then creates the embedding $h_v^{t+1}$ 
for vertex $v$ at time $t+1$
by combining its previous embedding  at time $t$ with the message at time $t+1$
obtained through aggregation. In both operations, it is assumed that the data
is transformed by a function, i.e., $M_t$ for aggregation and $U_t$ for 
update. In our implementation, $M_t$ is defined as a matrix multiplication 
between edge attribute linear transformations and node embeddings. This was defined in 
\cite{gilmer2017neural} as 
\[
M_t(h_v^t, h^t_w, e_{vw}) = A_{e_{vw}}h_w^t
\]

The update function $U_t$ is defined as a 
Gated Recurrent Unit (GRU) layer introduced in \cite{cho2014learning}.
Note that \cite{gilmer2017neural} states that the GRU will take as input $h_v^t$ and $h_v^0$ at every iteration, but we opt for the network from \cite{REISER2021100095} which instead takes in $h_v^t$ and $h_v^{t-1}$. This design choice was made after experimentally 
observing that the latter gives better results than the former.
Finally, we pass the output from the message passing layers into a 
multi-layer perceptron (MLP) to obtain the $Q$ and $F$ values.
In our implementation, we  use three message passing layers, i.e., the
maximum value for $t$ is three. Critical to the approach are the node
attributes to be used for $h$ at time $0$, as well as the edge attributes $e_{vw}$ prior to linearization.
These attributes will be defined in the remainder of this section.
In our use case, we create two different networks using this architecture, 
one for $Q$ and one for $F$\footnote{\label{fn:website}The complete GNN architecture as well as relevant training documentation can be found at \url{ucmercedrobotics.github.io/gnn-sop}.}. 
We have two separate networks since the MCTS training 
data is best captured by two different activation functions for $Q$ and $F$,
We select a linear activation 
function for the $Q$ value and a sigmoid for $F$. The reasoning is that $Q$ 
predicts a positive, but potentially unbounded scalar value while $F$ 
predicts a probability between 0 and 1.
We also hypothesized that the networks would train differently on both types of 
labels, so we opted to separate them. Indeed, preliminary experiments 
showed that our loss was reduced when separating the models completely 
instead of having one model with two output activation functions or two output MLPs. 
We selected mean squared error (MSE) as the loss function for both $Q$ and $F$ outputs, 
as the task at hand is regression.

\subsection{Attributes} \label{sec:attributes}
For each node in the graph, we introduce an eight dimensional vector of
attributes that will be used to define $h_v$ in Eq.~\eqref{eq:mpnn} for $t = 0$.
Each node has the following attributes:
    \begin{itemize}      
        \item a binary value indicating if the vertex has been visited;
        \item its $x$ and $y$ coordinates in the plane;
        \item its reward;
        \item the remaining budget;
        \item a binary value indicating if the node is the start vertex;
        \item a binary value indicating if the node is the end vertex;
        \item a binary value indicating if the node is the vertex where the 
        robot is currently positioned at.
    \end{itemize}
Our choice for these attributes is inspired by \cite{9109309}, where
a similar approach was used to solve the TSP. 
Accordingly, the first three attributes provide the model 
with
spatial awareness, which proved effective in TSP setting.
However, since SOPCC has more constraints than TSP, we added 
additional attributes such as reward and residual budget.
This is due to the fact that the nodes are not all uniformly important, and 
in general the agent will not be able to visit every node.
Intuitively, incorporating more spatial and temporal information helps the model
learn the greedy heuristic described in Section \ref{MCTS}. 
Indeed, as we will show in Section \ref{sec:ablation} through an ablation study,
these additions were critical to achieving performance comparable to previous methods.
For edge attributes $e_{vw}$, we set the single attribute to the Euclidean
distance between nodes $v$ and $w$. Note that the edge attributes do
not include any information about the length variability $d(e_{v,w})$.
This choice was based on preliminary tests, which showed no conclusive performance improvement from including this information.
Additionally, omitting it reduces the number of model parameters, enabling a smaller training
dataset and  reducing training time.

In selecting our model architecture, we drew from the literature reviewed in
Section~\ref{sec:sota} as well as a direct comparison using test cases.
We trained both a Message Passing Neural Network (MPNN) \cite{gilmer2017neural} and a Graph Attention Network (GAT) \cite{veličković2018graph}.
Our initial hypothesis was that the attention mechanism in the GAT might help
identify more valuable nodes by assigning higher weights to more promising neighbors.
However, in practice, the GAT underperformed slightly compared to the MPNN.
Specifically, the average reward achieved by the GAT was approximately 10\% lower
than that of our largest MPNN model, with only a marginal computational speedup
of about 50 ms over the entire MCTS pipeline, including inference.
This trade-off was not significant enough to justify selecting the GAT.
Therefore, we opted for the neural message passing model presented in
\cite{gilmer2017neural, REISER2021100095}, prioritizing its performance and training stability.
Notably, the implementation from \cite{REISER2021100095} includes a skip connection
from the initial node attributes to the post-message-passing layers.
We removed this skip layer to align more closely with the original design in
\cite{gilmer2017neural} and because it did not result in any performance improvements
in our experiments.

\subsection{Training} \label{training}
\begin{figure}[h]
\centering
\includegraphics[width=0.7\linewidth]{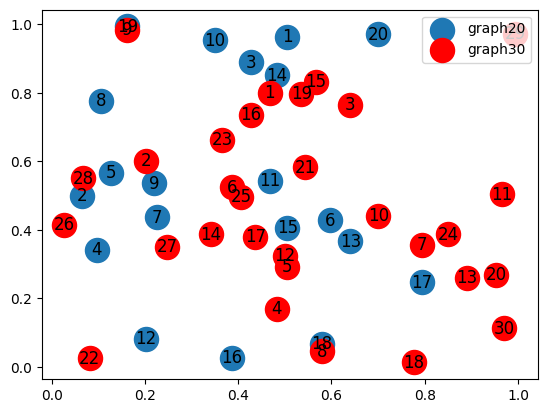}
\caption{Examples of randomly generated training graphs with twenty and thirty vertices.}
\label{fig:graph2030}
\end{figure}

To replace the rollout phase in MCTS with a trained model capable of estimating both utility ($Q$) and failure probability ($F$) without simulation, a necessary preliminary step is the generation of training data for the MPNN. To this end, we solved randomly generated instances of the SOPCC using the MCTS algorithm described in Section~\ref{MCTS}. These instances are defined over complete graphs, where rewards and $(x, y)$ coordinates are independently and uniformly sampled from the interval $[0, 1]$ (see Figure~\ref{fig:graph2030}), and edge lengths correspond to Euclidean distances. For training, we generated problem instances with 20, 30, and 40 vertices.

We trained all of our networks using a budget of $B = 2$ and a failure probability of $P_f = 0.1$. During training, travel costs along edges were sampled from an exponential distribution with a mean equal to the deterministic edge length (the same noise model used in our preliminary work \cite{CarpinTASE2024}). We observed that any network trained on more than 6,000 problem instance solutions appeared to converge, with the entire model size being 8.4 MB. Importantly, training data included only successful SOPCC solutions, i.e., cases where the MCTS algorithm produced routes that did not violate the budget constraint. We discuss the impact of these decisions in Section \ref{sec:results}.
\section{Results} \label{sec:results}

\subsection{Experiments}\label{sec:firstresults}
We experimentally evaluated the performance of our GNN-MCTS architecture in predicting the $Q$ and $F$ values for efficiently solving SOPCC instances. We compare our method against an exact approach based on mixed integer linear programming \cite{MILP} (referred to as MILP), as well as our previously proposed method in \cite{CarpinTASE2024}, referred to as MCTS-SOPCC. Additionally, for completeness, we include comparisons against the MCTS method described in Section \ref{MCTS} used during training and sharing similarities with MCTS-SOPCC but with minor implementation differences.

Our implementation is in Python 3.11.5, leveraging \cite{REISER2021100095} for GNN layers. Tests were conducted on a machine with an Intel Core i7-10700F (2.90 GHz), 64 GB RAM, and an Nvidia 1660 SUPER GPU. The MILP solutions were obtained using the commercial solver Gurobi.

The main objective is to assess the value network’s capability to learn heuristics and generalize across SOPCC instances of varying complexity. Compared to existing solvers, GNN-MCTS is expected to offer significant speed advantages and better scalability as problem size increases.

For fair comparison, both MCTS and MCTS-SOPCC use 100 rollouts ($S=100$) to estimate $Q$ and $F$ values by averaging results from 100 independent executions, consistent with \cite{CarpinTASE2024}. The search trees in these methods are expanded 350 times, also following \cite{CarpinTASE2024}. In all experiments, the failure probability threshold is set to $P_f = 0.1$.

    \begin{table*}   
    \center
        \begin{tabular}{||c|c|c|c|c||}
        \cline{2-5}
        \multicolumn{1}{c|}{} & \multicolumn{4}{c|}{GNN-MCTS Timing Splits} \\\cline{2-5}
        \multicolumn{1}{c|}{} & Select (s) & Expand (s) & Rollout (s)& \multicolumn{1}{c|}{Backpropagate (s)} \\
        \hline
        $\text{graph20}_{B=2}$ & $8.6 \times 10^{-7} \pm 1.2 \times 10^{-7}$ & $0.0002 \pm 1.7\times10^{-5}$ & $\mathbf{0.001 \pm 0.007}$ & $4.3 \times 10^{-5} \pm 6.7 \times 10^{-6}$\\
        $\text{graph30}_{B=2}$ & $1.0 \times 10^{-6} \pm 2.3 \times 10^{-7}$ & $0.0002 \pm 1.4\times10^{-5}$ & $\mathbf{0.001 \pm 0.006}$ & $5.2 \times 10^{-5} \pm 8.3 \times 10^{-6}$\\
        $\text{graph40}_{B=2}$ & $1.1 \times 10^{-6} \pm 1.2 \times 10^{-7}$ & $0.0002 \pm 2.1\times10^{-5}$ & $\mathbf{0.0009 \pm 0.005}$ & $6.3 \times 10^{-5} \pm 3.4 \times 10^{-6}$\\
        \hline
        \multicolumn{5}{c}{} \\
        \cline{2-5}
        \multicolumn{1}{c|}{} & \multicolumn{4}{c|}{MCTS Timing Splits} \\\cline{2-5}
        \multicolumn{1}{c|}{} & Select (s) & Expand (s) & Rollout (s)& \multicolumn{1}{c|}{Backpropagate (s)} \\
        \hline
        $\text{graph20}_{B=2}$ & $2.3 \times 10^{-6} \pm 8.7 \times 10^{-7}$ & $0.0002 \pm 2.6\times10^{-5}$ & $\mathbf{0.002 \pm 0.001}$ & $4.5 \times 10^{-5} \pm 1.2 \times 10^{-5}$\\
        $\text{graph30}_{B=2}$ & $2.3 \times 10^{-6} \pm 7.7 \times 10^{-7}$ & $0.0002 \pm 3.0\times10^{-5}$ & $\mathbf{0.011 \pm 0.008}$ & $5.3 \times 10^{-5} \pm 2.3 \times 10^{-5}$\\
        $\text{graph40}_{B=2}$ & $2.7 \times 10^{-6} \pm 7.8 \times 10^{-7}$ & $0.0003 \pm 3.7\times10^{-5}$ & $\mathbf{0.020 \pm 0.014}$ & $6.4 \times 10^{-5} \pm 1.5 \times 10^{-5}$\\
        \hline
        \end{tabular}
        \caption{Timing breakdown between MCTS with and without GNN value network over 100 trials.}
        \label{table:times}
    \center
        \vspace{-5mm}
    \end{table*}

    The results in Table \ref{table:times} show graphs with different numbers of vertices and their associated MCTS timing breakdowns. The most evident result from these experiments is how linearly the rollout timing scales with size, and even improves, on average, as graph size grows. This improvement is due to the single forward pass used to approximate utility and failure chance for all nodes in the graph at once. Note that we observe the standard deviation of average rollout times for the GNN-MCTS large due to the single forward pass, which is averaged out over time until the next required forward pass, after all children have been evaluated from a single node. As shown in Table \ref{table:results}, we see that the reward performance is very close while saving upwards of 1000\% in overall timing, with similar results as complexity scales.

 Timing aside, when solving SOPCC instances, one must consider both the reward collected and the failure rate. Accordingly, we define performance through two variables: the average reward obtained per solution, $R$, and the average rate of failure, $F$. Obviously, a method that collects high rewards but significantly exceeds the failure threshold $P_f$ is not practically valuable. Hence, a viable solution is one that remains within the constraints while maximizing cumulative rewards.
For this test, and those that follow, we reused the same test graphs supplied by \cite{CarpinTASE2024} for benchmarking, in addition to larger graphs.
Since we trained on instances with $P_f = 0.1$, we compare all methods using the same threshold.
As shown in Table \ref{table:results}, GNN-MCTS is, at times, more conservative and achieves lower failure rates $F$, while incurring only limited reward loss. However, in the case of larger, previously unseen graph sizes, we observe higher failure rates. We will further discuss the implications of network generalization in Section \ref{sec:generalization}.

  \begin{figure}[ht]
        \centering
        \includegraphics[angle=90, width=0.9\columnwidth]{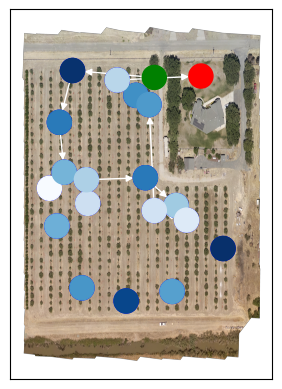}
        \caption{A test case of size 20 plotted over our test orchard. The white arrows indicate what our GNN-MCTS solved for this problem as the best path.
        }
        \label{fig:farm_graph}
    \end{figure}   
     
\begin{figure*}[htbp]
        \centering
        \begin{subfigure}[b]{0.3\textwidth}
            \centering
            \includegraphics[width=\textwidth]{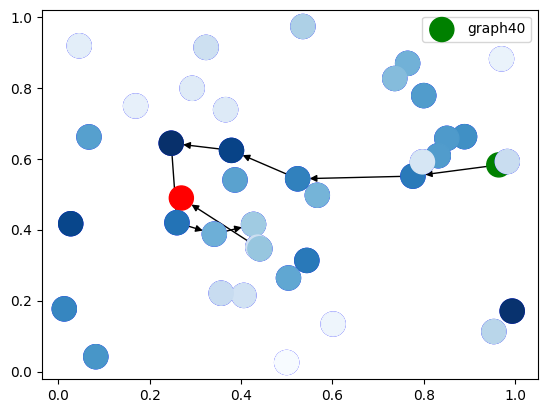}
            \caption{Graph40 GNN-MCTS solution}
            \label{fig:graph40_sol_gnn}
        \end{subfigure}
        \hfill
        \begin{subfigure}[b]{0.3\textwidth}
            \centering
            \includegraphics[width=\textwidth]{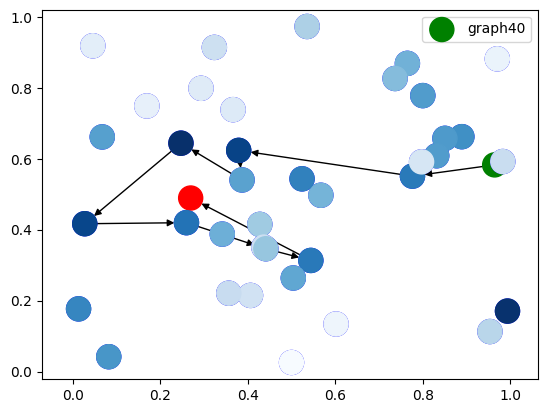}
            \caption{Graph40 MCTS solution}
            \label{fig:graph40_sol_mcts}
        \end{subfigure}
        \hfill
        \begin{subfigure}[b]{0.3\textwidth}
            \centering
            \includegraphics[width=\textwidth]{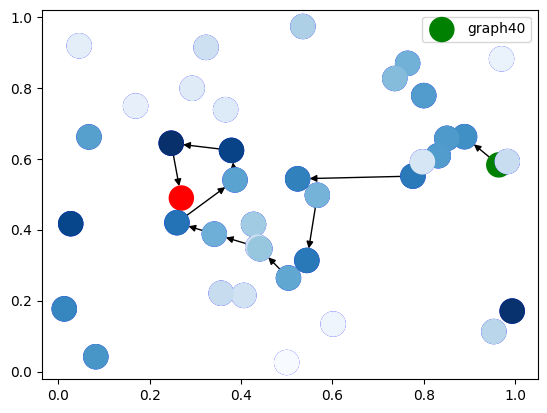}
            \caption{Graph40 MILP solution}
            \label{fig:graph40_sol_milp}
        \end{subfigure}
        \caption{Proposed solutions for three algorithms graph40 test described in Table \ref{table:results}. The green node is start; red is goal. The blue represents nodes of varying reward, darker being higher.}
        \label{fig:three_graphs_comp}
    \end{figure*}

In Table \ref{table:ratios}, we display the performance ratios between GNN-MCTS, the MCTS model used to generate its training data, and the MILP solver. This illustrates how remarkably fast and effective our model is compared to existing methods.
We stay within 93\% or better of our MCTS model, demonstrating that this architecture loses less than 10\% performance in these cases.
Even more impressively, in some cases we outperform the original model.
While performance against the MILP achieves 60\% or better in terms of reward, we expect improvements with optimal training data—which is significantly more difficult and time-consuming to generate.
Figure \ref{fig:three_graphs_comp} compares the solutions produced by three of the evaluated algorithms for an SOPCC instance with 40 vertices.
    
    \begin{table*}[t]
    \begin{center}
        \begin{tabular}{||c|c|c|c|c|c|c|c|c|c|c|c|c|c||}
        \cline{2-14}
         \multicolumn{1}{c|}{} & $P_f$ & \multicolumn{3}{c|}{GNN-MCTS} & \multicolumn{3}{c|}{MCTS} & \multicolumn{3}{c|}{MCTS-SOPCC} &  \multicolumn{3}{c||}{MILP}\\\cline{3-14}
         \multicolumn{1}{c|}{} & & Reward & Time (s) & $F$ & Reward & Time (s) & $F$ & Reward & Time (s) & $F$ & Reward & Time (s) & $F$ \\
         \cline{2-14}\hline
         $\text{graph20}_{B=2}$ & 0.1 & 2.762 & 0.226 & 4\% & 2.967 & 5.092 & 10\% & 3.49 & 10.99 & 11\% & 3.414 & 316.16 & 12\%\\
         $\text{graph30}_{B=2}$ & 0.1 & 5.675 & 0.442 & 10\% & 5.517 & 43.969 & 12\% & 6.433 & 23.117 & 7\% & 6.973 & 9.949 &10\%\\
         $\text{graph40}_{B=2}$ & 0.1 & 5.523 & 0.447 & 13\% & 5.75 & 70.712 & 11\% & 8.052 & 43.698 & 10\% & 8.843 & 457.13 & 10\%\\         
         \hline \hline
$\text{graph20}_{B=3}${\color{green}$^\&$} & 0.1 & 4.27 & 0.329 & 4\% & 4.207 & 7.003 & 7\% & 5.481 & 26.64 & 13\% & 5.404 & 601.078 & 14\%\\
         $\text{graph30}_{B=3}${\color{green}$^\&$} & 0.1 & 6.519 & 0.483 & 7\% & 6.804 & 59.63 & 14\% & 7.934 & 47.848 & 11\% & 8.779 & 317.093 &10\%\\
         $\text{graph40}_{B=3}${\color{green}$^\&$} & 0.1 & 7.106 & 0.555 & 1\% & 7.343 & 99.019 & 13\% & 10.434 & 77.774 & 12\% & 11.548 & 600.609 & 14\%\\  
         
         \hline
         \hline
         $\text{graph50}_{B=2}${\color{blue}$^*$} & 0.1 & 6.952 & 0.533 & 18\%{\color{red}$^!$} & 6.422 & 45.935 & 10\% & 8.394 & 26.06 & 4\% & 9.019 & 600.79 & 13\%\\  
         $\text{graph60}_{B=2}${\color{blue}$^*$} & 0.1 & 8.889 & 0.756 & 30\%{\color{red}$^!$} & 7.429 & 148.93 & 11\% & 10.29 & 31.031 & 10\% &10.602 & 601.09 & 10\%\\    
         $\text{graph70}_{B=2}${\color{blue}$^*$} & 0.1 & 8.462 & 0.778 & 40\%{\color{red}$^!$} & 8.381 & 205.451 & 12\% & 11.721 & 31.222 & 11\% & 12.35 & 601.459 & 10\%\\   
        \hline
        \end{tabular}
        \caption{
        Overall results from benchmarks used in \cite{CarpinTASE2024} plus even larger graphs averaged over 100 trials. Top rows are for results against previously seen budgets. {\color{blue}$^*$} signify tests run with a model trained against size 40 graphs. {\color{green}$^\&$} are for results against unseen budgets using same sized model. {\color{red}$^!$} is further explained in \ref{sec:generalization}.
        }
        \label{table:results}
    \end{center}
    \end{table*}

    \begin{table}   
    \center
        \begin{tabular}{||c|c|c|c|c||}
        \cline{2-5}
        \multicolumn{1}{c|}{} & \multicolumn{2}{c|}{GNN-MCTS / MILP} & \multicolumn{2}{c||}{GNN-MCTS / MCTS}\\\cline{2-5}
        \multicolumn{1}{c|}{} & $R$ Ratio & Time Ratio & $R$ Ratio & Time Ratio \\
        \hline
        $\text{graph20}_{B=2}$ & 0.809 & 1428.57 & 0.931 & 22.73\\
        $\text{graph30}_{B=2}$ & 0.814 & 22.73 & 1.029 & 100.00\\
        $\text{graph40}_{B=2}$ & 0.625 & 1111.11 & 0.961 & 158.73\\
        \hline
        \hline
        $\text{graph20}_{B=3}$ & 0.79 & 2000.00 & 1.015 & 21.28 \\
        $\text{graph30}_{B=3}$ & 0.743 & 666.67 & 0.958 & 123.46\\
        $\text{graph40}_{B=3}$ & 0.615 & 1075.27 & 0.968 & 178.57\\
        \hline
        \end{tabular}
        \caption{Ratios between our solver and baseline solvers for reward and time averaged over 100 trials. See Table \ref{table:unseen_sizes} for unseen graph sizes.}
        \label{table:ratios}
    \center
        \vspace{-3mm}

    \end{table}

It is also interesting to explore how the solver performs as $P_f$ varies.
While in most cases, model performance characterized by $P_f$ was as expected, a limitation of our implementation is that the activation function modeling $F$ is a sigmoid, which does not fully capture failure as a function.

Ultimately, we achieved real-time approximation results in a stochastic environment. Figure \ref{fig:farm_graph} shows the solution produced by GNN-MCTS while solving an instance of SOPCC associated with a sampling problem in precision agriculture. The blue dots represent locations to visit, with the color intensity indicating the reward (darker means higher reward). The robot starts at the green vertex and must end at the red vertex before running out of budget.

\subsection{Ablation Study}\label{sec:ablation}
Previous literature has shown that including spatial information in vertex attributes is essential for evaluating the effectiveness of GNN-based methods on problems like the TSP \cite{DBLP:journals/corr/DaiKZDS17}. However, because SOPCC solutions are also influenced by rewards, remaining budget, and the start and goal locations, as discussed in Section \ref{sec:attributes},
we extended the attribute set accordingly in our implementation. This subsection presents an ablation study to assess the impact of our novel temporal and spatial attribute additions, specifically the one-hot encoding of the currently visited vertex and the start/goal vertices.

    \begin{figure}[htb]
        \centering
        \includegraphics[width=1\linewidth]{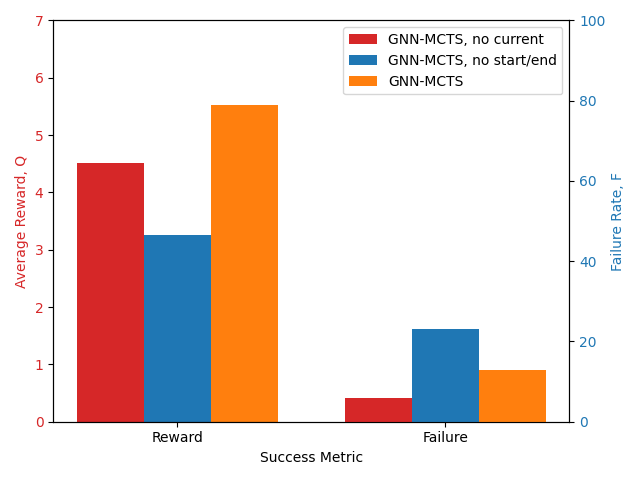}
        \caption{Ablation study substantiating that both temporal and spatial node attribute additions are needed for the value network to perform effectively 
        as removing either one implies decrease in reward collected (left) or increase
        in failure rate (right).
        Data averaged over 100 trials with $P_f = 0.1$.}
        \label{fig:ablation}
            \vspace{-5mm}
    \end{figure}   
    
In Figure \ref{fig:ablation}, we see that removing any of the additional attributes results in a decrease in average reward. In the case of removing the start/end vertex attribute, we observed nearly a 50\% loss in reward. However, the most significant impact is on the failure rate, which exceeds the $P_f$ constraint and renders the solution infeasible. The same figure also shows that omitting the current vertex attribute does not lead to a violation of $P_f$, but the average total reward is worse than that of the original model. While this solution is more conservative, achieving a lower $F$ rate, our original model maintains compliance with $P_f$ while collecting, on average, more reward.

For node-level classification using GNNs, these findings are relatively significant: GNNs can capture temporal and spatial relationships relevant to a given problem. Rather than embedding this information elsewhere in the algorithm, we can instead task the GNN with learning explicitly defined spatiotemporal data.
    
\subsection{Generalization} \label{sec:generalization}

As with any learning-based approach, the quality of the solution depends heavily on the nature of the training dataset. In our work, with the explicit goal of assessing the model’s ability to learn and generalize, we trained using a fixed failure probability $P_f = 0.1$ and budget $B = 2$, employing the same underlying graph topology, i.e., complete graphs with vertices uniformly distributed in the unit square. Given these premises, it is essential to evaluate how well the proposed architecture generalizes to instances characterized by new parameters. We investigated generalization from two perspectives: varying graph sizes and previously unseen budgets. The ability to generalize across these parameters is important not only for robustness but also to accelerate training, particularly with respect to graph size, since larger instances require significantly more time to generate solutions using the MCTS algorithm employed during training. As noted in Section \ref{training}, our training data consisted solely of graphs with 20, 30, and 40 vertices.

For generalization with respect to size, we observed in Table \ref{table:results} that within a certain size range, the model delivers equally competitive performance. For example, we pushed our model to the high end of what is seen in the SOPCC literature (e.g., up to 70 vertices) and found that generalization is quite effective in terms of reward performance. However, due to the sigmoid issue described earlier, for larger problem instances the model requires tuning of the $P_f$ value. Specifically, as shown in Table \ref{table:results}, when feeding the model a desired $P_f$ value for unseen problem instances, the solutions can have failure probabilities that significantly exceed the limit. This issue can be remedied by providing the model with a $P_f$ value smaller than the desired one. Note that this hyperparameter tuning process is also necessary when using the exact method \cite{MILP} and is not a limitation unique to our method. Table \ref{table:unseen_sizes} demonstrates how, by adjusting the $P_f$ value in unseen problem instances, GNN-MCTS successfully meets the originally desired failure rate.

    \begin{table}   
    \center
        \begin{tabular}{||c|c|c|c|c||}
        \cline{2-5}
        \multicolumn{1}{c|}{} & $P_f$ & $R$ & MCTS $R$ Ratio & $F$ \\
        \hline
        $\text{graph50}_{B=2}$ &0.075 & 6.171 & 0.961 & 6\%\\
        $\text{graph60}_{B=2}$ & 0.06 & 7.531 & 1.014 & 8\%\\
        $\text{graph70}_{B=2}$ & 0.075 & 8.007 & 0.955 & 7\%\\
        \hline
        \end{tabular}
        \caption{When hyperparameterizing $P_f$ on unseen graph sizes, we see the model performance improve to near ground truth, presumably due to the sigmoid activation function.}
        \label{table:unseen_sizes}
    \center
        \vspace{-7mm}
    \end{table}

   In the case of budget generalization, as shown in Table \ref{table:results}, our method was able to maintain similar time and performance ratios, detailed in Table \ref{table:ratios}, for unseen budgets. Given that solving a SOP with a lower budget is more difficult, it makes sense that providing more budget simplifies the task for the model. However, we observe our algorithm exhibiting noticeably lower failure rates, $F$, without sacrificing reward, $R$, implying once again that our method is more conservative without a significant performance impact.

\section{Conclusions and Future Work}
\label{sec:conclusions}

In this paper, we have presented a novel solution to the SOPCC by combining MCTS with a message passing graph neural network. This is achieved by training a neural network to replace the computationally expensive rollout phase in MCTS. Building on the key idea introduced in \cite{CarpinTASE2024}, the trained model simultaneously predicts both the expected collected reward, $Q$, and the predicted failure probability, $F$. This capability was not found in previous works employing value networks with reinforcement learning. Another advantage of our methodology is that, during the expansion phase, a single forward pass of the network predicts values for all child nodes simultaneously, rather than evaluating them one by one. This enables the algorithm to efficiently explore a larger portion of the search space.

Our method was trained using SOPCC solutions generated by a slightly modified version of the algorithm presented in \cite{CarpinTASE2024}. Extensive simulations compared this new approach against the training method, our previous MCTS algorithm \cite{CarpinTASE2024}, and an exact MILP-based solution. The results demonstrate that our proposed method is orders of magnitude faster while satisfying the failure constraint and preserving a significant portion of the collected reward, though some hyperparameter tuning may be required in certain cases. Our ablation study further confirms that the selected vertex attributes for the embeddings are critical to the method’s success.

There are numerous directions for current and future work. First, it would be interesting to train the model using solutions generated by different heuristics or even an ensemble of heuristics to assess whether this improves overall performance, especially in generalizing to larger problem instances. We also plan to explore applying this model, either in centralized or distributed settings, to multi-robot problems involving communication. Additionally, investigating alternative vertex or edge attributes may enhance the predictive capabilities of the network. Similarly, modifications to the network’s final layers, such as improved activation functions, could yield better results. Finally, we are currently implementing the planner in ROS 2 and integrating it into our autonomous navigation stack for precision agriculture, assessing its real-time applicability to complex data collection problems in the field.

\bibliographystyle{plain}
\bibliography{report.bib}

\end{document}